%% file: root.tex
\documentclass[letterpaper, 10 pt, conference]{ieeeconf}

\IEEEoverridecommandlockouts 
\usepackage{amsmath}
\usepackage{amssymb}
\usepackage{amsfonts}

\usepackage{microtype} %
\usepackage{cite}

\usepackage{tikz}
\usetikzlibrary{arrows.meta, positioning, fit, calc, backgrounds, shapes.geometric}
\usepackage{acro}
\input{acronyms.tex}

\usepackage[hidelinks]{hyperref}
\usepackage[capitalise, nameinlink]{cleveref}

\crefformat{equation}{#2(#1)#3}
\crefrangeformat{equation}{#3(#1)#4 to~#5(#2)#6}
\crefmultiformat{equation}{#2(#1)#3}{ and~#2(#1)#3}{, #2(#1)#3}{ and~#2(#1)#3}

\Crefformat{equation}{Equation~#2(#1)#3}
\Crefrangeformat{equation}{Equations~#3(#1)#4 to~#5(#2)#6}
\Crefmultiformat{equation}{Equations~#2(#1)#3}{ and~#2(#1)#3}{, #2(#1)#3}{ and~#2(#1)#3}

\title{\LARGE \bf
Parallel-in-Time Nonlinear Optimal Control via GPU-native Sequential Convex Programming
}

\author{Yilin Zou$^{1}$, Zhong Zhang$^{1,2}$, Maxime Robic$^{3}$ and Fanghua Jiang$^{1}$%
\thanks{*This work was supported by the National Natural Science Foundation of China (Grant Nos. 12472355 and 12532018).}%
\thanks{$^{1}$Yilin Zou is a Ph.D. candidate, Zhong Zhang is with, and Fanghua Jiang is a Professor with the School of Aerospace Engineering, Tsinghua University, 100084 Beijing, China.
        {\tt\small zouyl22@mails.tsinghua.edu.cn, jiangfh@tsinghua.edu.cn} (Corresponding Author: Zhong Zhang and Fanghua Jiang)}%
\thanks{$^{2}$Zhong Zhang is also a Marie Sk{\l}odowska-Curie Postdoctoral Fellow with the Department of Aerospace Science and Technology, Politecnico di Milano, 20156 Milan, Italy.
        {\tt\small zhong.zhang@polimi.it} }%
\thanks{$^{3}$Maxime Robic is a Marie Sk{\l}odowska-Curie Postdoctoral Fellow with the Department of Aerospace Science and Technology, Politecnico di Milano, 20156 Milan, Italy.
        {\tt\small maxime.robic@polimi.it}}%
} 

\usepackage{bm} %
\newcommand{\mathbfit}[1]{\bm{#1}} %
\usepackage{bbm} 
\usepackage{float}

\begin{document}
\maketitle

\thispagestyle{empty}
\pagestyle{empty}

\begin{abstract}
Real-time solution of nonlinear optimal control problems remains challenging on embedded robotic hardware, where conventional solvers often rely on global sparse linear algebra or sequential recursions that are difficult to map efficiently to massively parallel processors. This paper presents \textit{ucenter}, a GPU-native Sequential Convex Programming (SCP) framework for nonlinear optimal control. At each SCP iteration, nonlinear dynamics are linearized around a nominal trajectory, and the resulting convexified subproblem is solved by a consensus Alternating Direction Method of Multipliers (ADMM) scheme. The temporal splitting replaces global sparse Karush--Kuhn--Tucker factorizations with independent per-node dense solves, closed-form dynamic consistency updates, and analytical projections onto convex constraint sets. Both the outer SCP loop and the inner ADMM subproblem are executed entirely on the GPU, enabling efficient optimization.

The proposed solver is evaluated on quadrotor obstacle avoidance and Mars powered descent problems using an NVIDIA Jetson AGX Orin edge platform. Benchmarking against a CPU-parallel iLQR baseline in randomized environments reveals that the GPU implementation achieves over 100 Hz batched planning throughput, a 4.1x speedup, and a 51\% reduction in energy consumption, while consistently maintaining low nonlinear dynamics defects. The framework exposes reusable GPU-parallel optimization primitives that can be specialized to a wide variety of complex nonlinear optimal control settings, as demonstrated by the scenario-based robust MPC and batched Monte Carlo generation tasks.
\end{abstract}

\input{illustration}

\section{Introduction}
\label{sec:intro}

Real-time trajectory optimization plays a crucial role in modern autonomous systems, enabling agile quadrotors to execute aggressive maneuvers~\cite{foehn2021time}, reusable launch vehicles to perform pinpoint rocket-powered landings~\cite{acikmese2007lossless, szmuk2020successive}, and high-DOF manipulators to operate near their physical limits~\cite{sundaralingam2023curobo}. As robotic missions become increasingly complex, the demand for solving large-scale, nonconvex \acp{ocp} in real time has intensified. Despite significant progress, many conventional solvers remain organized around CPU-oriented sparse linear algebra, sequential recursions, or limited forms of parallelism. This creates a bottleneck for exploiting modern massively parallel hardware such as GPUs.

In continuous-time optimal control, trajectory optimization methods are commonly discussed in terms of indirect and direct methods. Indirect methods derive necessary optimality conditions from Pontryagin's Maximum Principle, but the resulting boundary value formulations are often sensitive to initialization and less convenient for general path constraints. This paper focuses on direct methods, which discretize the optimal control problem into a finite-dimensional optimization problem and are widely used in practical constrained trajectory optimization.

A useful implementation-level distinction among direct methods is whether the state trajectory is recovered mainly by forward simulation or treated explicitly as an optimization variable together with the control sequence. In classical single-shooting formulations, \ac{ddp} and its variants, including iLQR~\cite{liIterativeLinearQuadratic2004}, primarily optimize the control sequence while the state trajectory is recovered through forward rollout. These methods are computationally efficient and effective for smooth unconstrained or soft-constrained problems. Recent work has explored temporal and hardware parallelism for \ac{ddp}/iLQR-type methods, including a GPU implementation of a multiple-shooting \ac{ddp} variant~\cite{plancher2018performance}, associative-scan formulations for dynamic programming and linear-quadratic control~\cite{sarkka2021temporal}, and a primal-dual GPU-iLQR method that uses parallel associative scans to solve the primal-dual \ac{kkt} system~\cite{Amatucci_2026}. Handling hard state and control constraints in \ac{ddp}/iLQR-style methods typically requires additional mechanisms such as penalties, barriers, augmented Lagrangian terms, or specialized constrained variants~\cite{howell2019altro}. These extensions improve applicability to constrained problems, but they also add algorithmic components beyond the basic Riccati-style recursion.

In contrast, direct transcription methods treat both states and controls as optimization variables. Examples include collocation and pseudospectral discretizations, which explicitly impose defect, path, and boundary constraints within the resulting finite-dimensional problem. In trajectory optimization, \ac{scp} is often instantiated as successive convexification, where nonlinear dynamics and nonconvex constraints are repeatedly linearized or convexified around the current trajectory and solved as a sequence of convex subproblems~\cite{mao2016successive,szmuk2020successive}. While these formulations are flexible, direct transcriptions often lead to large sparse nonlinear programs, and \ac{scp} methods lead to sequences of sparse convex subproblems. Robust implementations of these problems commonly rely on centralized sparse linear algebra, for example through general-purpose \ac{nlp} solvers such as IPOPT~\cite{wachterImplementationInteriorpointFilter2006} or sparse convex optimization backends. Direct sparse factorization of large, irregular \ac{kkt} systems can involve nonuniform memory access and synchronization patterns that are less directly aligned with \ac{simt} execution~\cite{nocedal2006numerical,pacaud2024gpuaccelerateddynamicnonlinearoptimization}. This motivates solver structures that expose more uniform, fine-grained parallelism across the trajectory horizon.

Recent work has leveraged GPU architectures across different trajectory optimization and \ac{mpc} formulations. Beyond Riccati/iLQR parallelization, Adabag et al. introduced MPCGPU, a GPU-accelerated nonlinear \ac{mpc} solver based on a custom preconditioned conjugate gradient method for the sparse linear systems arising in direct trajectory optimization~\cite{adabag2024mpcgpu}. Chari et al. developed a GPU-accelerated \ac{scp} framework for 6-DoF powered descent guidance, using a proportional-integral projected gradient solver to conduct large-scale Monte Carlo simulations efficiently~\cite{chari2024fast}. Massively parallel motion generation libraries such as cuRobo~\cite{sundaralingam2023curobo} further demonstrate the value of many parallel seeds and trajectory candidates on GPUs for collision-free motion generation. These works show that GPU acceleration can benefit a wide range of optimal control and motion planning formulations, while leaving room for reusable GPU-native frameworks that combine direct transcription with parallel execution based on operator splitting.

To address this gap, this paper presents \textit{ucenter}, a GPU-native \ac{scp}--\ac{admm} framework for nonlinear path-constrained optimal control (see \cref{fig:method_overview}). Instead of relying on global sparse \ac{kkt} systems at each \ac{scp} iteration, \textit{ucenter} exploits time-node parallelism through consensus-based operator splitting \cite{boyd2011distributed,odonoghue2013splitting,morgan2016model}. The trajectory is decoupled into local node variables linked by dynamic consistency constraints. The resulting \ac{admm} scheme consists of independent dense solves per timestep, closed-form dynamic consistency updates, and analytical projections for state and control constraints. By executing these operations, alongside linearization and trajectory updates, concurrently on the GPU, the framework keeps the entire optimization loop on the device and minimizes CPU-GPU synchronization.

These highly parallel computational primitives extend seamlessly beyond single-trajectory planning. Because the framework exposes generic operator-splitting structures, it can be readily specialized to multi-trajectory optimal control settings. For instance, batched trajectory optimization can rapidly generate large collections of expert demonstrations for data-driven policies \cite{izzo2024optimality}, overcoming the computational bottleneck of repeated CPU-based solves \cite{carius2020mpc,levine2013guided}. Similarly, scenario-based robust or stochastic \ac{mpc} can be formulated over an ensemble of trajectories representing different uncertainty realizations \cite{campi2009scenario,mesbah2016stochastic}. In these scenario-coupled problems, consensus constraints efficiently enforce shared decision variables, such as the initial physical control action. \textit{ucenter} provides a unified architecture for a wide range of optimal control tasks, including single-trajectory planning, batched Monte Carlo generation, and scenario-based \ac{mpc}.

In summary, the main contributions of this paper are as follows:
\begin{enumerate}
    \item We present \textit{ucenter}, a GPU-native \ac{scp} framework for nonlinear path-constrained optimal control. The implementation keeps linearization, subproblem solves, and trajectory updates resident on the GPU, reducing CPU--GPU synchronization and supporting high-throughput execution on embedded hardware.

    \item We develop an \ac{admm}-based temporal splitting formulation for the convexified \ac{scp} subproblems. This formulation replaces global sparse \ac{kkt} factorizations with local node-wise updates, dynamic-consistency consensus steps, and analytical projections onto state and control constraint sets.

    \item We show that the same computational structure extends naturally to parallel multi-trajectory optimal control. This enables, among other applications, batched randomized trajectory generation and scenario-coupled \ac{mpc} with shared decision variables such as the first applied control.

    \item We validate the framework on quadrotor obstacle-avoidance and Mars powered-descent problems using an NVIDIA Jetson AGX Orin edge platform. The experiments include a comparison with a CPU-parallel iLQR baseline, hardware-utilization and active-energy measurements, randomized batch optimization, and scenario-based robust \ac{mpc}. We report nonlinear dynamics defects, constraint violations, boundary-condition errors, and timing/energy results to characterize practical embedded performance.
\end{enumerate}

The proposed framework is packaged as a reusable Python library. The source code will be made publicly available upon publication.

\section{Problem Formulation}
\label{sec:problem_formulation}

We consider the optimal trajectory generation for a nonlinear dynamical system over a discretized time horizon $t_0, \dots, t_N$ with step size $\Delta t$. Let $\mathbfit{x} = \{x_0, \dots, x_N\}$ and $\mathbfit{u} = \{u_0, \dots, u_{N-1}\}$ denote the state and control sequences, where $x_i \in \mathbb{R}^{n_x}$ and $u_i \in \mathbb{R}^{n_u}$. The discrete-time \ac{ocp} is formulated as:
\begin{subequations} \label{eq:ocp}
\begin{align}
    \min_{\mathbfit{x}, \mathbfit{u}} \quad & \phi(x_N) + \sum_{i=0}^{N-1} \ell(x_i, u_i) \label{eq:cost} \\
    \text{s.t.} \quad & x_{i+1} = f(x_i, u_i), \quad i = 0, \dots, N-1 \label{eq:dynamics} \\
    & x_0 = x_{\text{init}}, \quad S x_N = x_{\text{target}} \label{eq:boundary} \\
    & x_i \in \mathcal{X}_i, \quad u_i \in \mathcal{U}_i \label{eq:constraints}
\end{align}
\end{subequations}
Here, $\phi(\cdot)$ and $\ell(\cdot)$ define the terminal and running costs. \cref{eq:dynamics} represents the nonlinear system dynamics integrated over $\Delta t$ (e.g., via 4th-order Runge-Kutta). The boundary conditions \cref{eq:boundary} enforce the initial state $x_{\text{init}}$ and a partial terminal state $x_{\text{target}}$, mapped by a constant selection matrix $S$ to constrain only specific components. 

Finally, \cref{eq:constraints} restricts the states and controls to admissible closed convex sets $\mathcal{X}_i$ and $\mathcal{U}_i$. We assume these sets are ``prox-friendly,'' meaning they admit computationally cheap analytical projections. Typical examples include box constraints (e.g., joint limits handled via element-wise clamping) and norm constraints (e.g., maximum thrust bounds handled via vector scaling). Because these projection operations require no iterative sub-routines, they are ideally suited for massive GPU parallelization.

\section{Methodology}
\label{sec:method}

We propose a two-layer hierarchical framework, illustrated in \cref{fig:method_overview}, to solve the non-convex optimal control problem \eqref{eq:ocp}. The outer layer employs \ac{scp} to handle nonlinearities by iteratively constructing local quadratic approximations. The inner layer utilizes a massively parallel, consensus-based \ac{admm} solver to resolve the resulting large-scale \ac{qp} subproblems.

\subsection{Outer Loop: The \ac{scp} Interface}
\label{subsec:scp}

The primary role of the \ac{scp} layer is to convert the nonlinear dynamics and non-convex costs into a sequence of affine equality constraints and convex quadratic objectives. Let $(\bar{\mathbfit{x}}^k, \bar{\mathbfit{u}}^k)$ denote the nominal trajectory at iteration $k$. By performing a first-order Taylor expansion of the dynamics and a second-order expansion of the cost function around this nominal trajectory, we obtain the following convex \ac{qp} subproblem:

\begin{subequations} \label{eq:qp_subproblem}
\begin{align}
    \min_{\mathbfit{x}, \mathbfit{u}} \quad & \sum_{i=0}^{N} \left( \frac{1}{2} x_i^T Q_i x_i + q_i^T x_i \right) \nonumber \\
    &+ \sum_{i=0}^{N-1} \left( \frac{1}{2} u_i^T R_i u_i + x_i^T M_i u_i + r_i^T u_i \right) \label{eq:qp_cost} \\
    \text{s.t.} \quad & x_{i+1} = A_i x_i + B_i u_i + d_i, \quad i = 0, \dots, N-1 \label{eq:qp_dynamics} \\
    & x_i \in \mathcal{X}_i \cap \mathcal{X}_i^k, \quad u_i \in \mathcal{U}_i \cap \mathcal{U}_i^k \label{eq:qp_constraints}
\end{align}
\end{subequations}
Here, the matrices $A_i$ and $B_i$ are the Jacobians of the discretized dynamics $f$ evaluated at $(\bar{x}_i^k, \bar{u}_i^k)$, and $d_i$ represents the linearization residual. The matrices $Q_i, R_i, M_i$ and vectors $q_i, r_i$ form the local quadratic approximation of the running cost $\ell(\cdot)$, with $Q_N$ and $q_N$ specifically representing the Hessian and gradient of the terminal cost $\phi(\cdot)$. The sets $\mathcal{X}_i^k$ and $\mathcal{U}_i^k$ denote the trust-region bounds (typically $\ell_\infty$-norm box constraints) imposed to ensure the validity of the local linearizations.

Crucially, the evaluation of the nonlinear dynamics, the cost function, and their respective derivatives at each time step $i$ are entirely independent operations. This structure allows the outer \ac{scp} loop to compute the entire linear-quadratic approximation simultaneously by mapping each collocation node to a separate GPU thread block.

Since the focus of this work is the highly parallelized solution of the inner problem \eqref{eq:qp_subproblem}, we omit standard trust-region update logic and refer readers to literature such as \cite{nocedal2006numerical,mao2019successiveconvexificationsuperlinearlyconvergent} for a comprehensive treatment.

\subsection{Inner Loop: Parallel Consensus \ac{admm}}
\label{subsec:admm}

Solving the \ac{qp} subproblem \eqref{eq:qp_subproblem} using conventional solvers (e.g., \ac{sqp} or \ac{ipm}) is computationally expensive due to the temporal coupling introduced by the dynamics constraints. To expose parallelism, we reformulate the problem using the \ac{admm} framework with a specific variable splitting strategy.

\subsubsection{Variable Splitting and Reformulation}
To decouple the optimization horizon, we introduce three distinct sets of variables (mapped to the layers shown in \cref{fig:method_overview}), each handling a specific aspect of the control problem:
\begin{enumerate}
    \item \textbf{Physical Variables} $(x, u)$: These are the primal variables responsible for minimizing the local quadratic cost and satisfying the linearized dynamics. In the split form, they constitute an unconstrained quadratic program.
    \item \textbf{Dynamic Auxiliary Variables} $(z)$: Defined such that $x_i = z_i$. These variables decouple the temporal dependency between time steps $i$ and $i+1$, allowing the physical variables to be updated independently.
    \item \textbf{Geometric Mirror Variables} $(\hat{x}, \hat{u})$: Defined such that $x_i = \hat{x}_i$ and $u_i = \hat{u}_i$. These variables handle all hard inequality constraints (such as trust regions and actuation limits) via indicator functions, which correspond to proximal projection operations \cite{parikh2014proximal}.
\end{enumerate}

\subsubsection{The Augmented Lagrangian}
We formulate the problem as minimizing the Augmented Lagrangian $\mathcal{L}_{\rho}$. Hard constraints are converted into indicator functions $I_{\mathcal{C}}(\cdot)$, while consistency constraints are enforced via linear dual terms and quadratic penalties:
\begin{equation}
\label{eq:augmented_lagrangian}
\begin{aligned}
    \mathcal{L}_{\rho} = & \sum_{i=0}^{N} l_i(x_i, u_i) 
    + \sum_{i=0}^{N} I_{\mathcal{X}_i \cap \mathcal{X}_i^k}(\hat{x}_i) 
    + \sum_{i=0}^{N-1} I_{\mathcal{U}_i \cap \mathcal{U}_i^k}(\hat{u}_i) \\
    & + \sum_{i=1}^{N} \left( \lambda_i^T(x_i - z_i) + \frac{\rho_{\text{eq}}}{2}\|x_i - z_i\|_2^2 \right) \\
    & + \sum_{i=0}^{N-1} \left( \mu_{i+1}^T(z_{i+1} - d_{\text{dyn}, i}) + \frac{\rho_{\text{dyn}}}{2}\|z_{i+1} - d_{\text{dyn}, i}\|_2^2 \right) \\
    & + \sum_{i=0}^{N} \left( \nu_{x,i}^T(x_i - \hat{x}_i) + \frac{\rho_{\text{geo}}}{2}\|x_i - \hat{x}_i\|_2^2 \right) \\
    & + \sum_{i=0}^{N-1} \left( \nu_{u,i}^T(u_i - \hat{u}_i) + \frac{\rho_{\text{geo}}}{2}\|u_i - \hat{u}_i\|_2^2 \right)
\end{aligned}
\end{equation}
where $l_i(x_i, u_i)$ is the local quadratic cost defined in \eqref{eq:qp_cost}, and $d_{\text{dyn}, i} = A_i x_i + B_i u_i + d_i$ represents the linearized dynamic propagation. The vectors $\lambda, \mu, \nu$ are the dual multipliers, and $\rho_{\text{eq}}, \rho_{\text{dyn}}, \rho_{\text{geo}}$ are the scalar penalty parameters for state consistency, dynamic propagation, and geometric constraints, respectively.

\subsubsection{The \ac{admm} Iteration Steps}
The algorithm alternates between minimizing $\mathcal{L}_{\rho}$ with respect to each variable block. Due to the temporal splitting, these steps can be executed in parallel or resolved via closed-form solutions across the trajectory nodes.

\textbf{Step 1: Physical Layer Update $(x, u)$.}
We minimize the terms in $\mathcal{L}_{\rho}$ involving $x_i$ and $u_i$, treating all other variables as constants. Ignoring indicator functions results in an unconstrained quadratic minimization. For each time step $i$, this requires solving the linear system $H_i \xi_i = g_i$, where $\xi_i = [x_i^T, u_i^T]^T$.
The Hessian matrix $H_i$ aggregates curvature from the original cost \eqref{eq:qp_cost} and the quadratic penalty terms:
\begin{equation}
    H_i = \begin{bmatrix}
    H_{11,i} & H_{12,i} \\
    H_{21,i} & H_{22,i}
    \end{bmatrix}
\end{equation}
where the block components are defined as:
\begin{align*}
    H_{11,i} &= Q_i + (\rho_{\text{geo}} + \rho_{\text{eq}}\mathbbm{1}_{i>0})I + \rho_{\text{dyn}}A_i^T A_i \\
    H_{12,i} &= M_i + \rho_{\text{dyn}}A_i^T B_i \\
    H_{21,i} &= M_i^T + \rho_{\text{dyn}}B_i^T A_i \\
    H_{22,i} &= R_i + \rho_{\text{geo}}I + \rho_{\text{dyn}}B_i^T B_i
\end{align*}
Crucially, $H_i$ is strictly positive definite due to the $\rho$ regularization terms. This guarantees the mathematical stability of standard, unpivoted Cholesky factorization. By avoiding numerical pivoting, this step completely circumvents branching and thread divergence, making it well suited for \ac{simt} GPU execution. Furthermore, since $H_i$ is constant within an \ac{scp} iteration, its factorization can be cached, leaving only highly efficient forward-backward substitutions for the inner \ac{admm} loop.

\textbf{Step 2: Dynamic Layer Update $(z)$.}
We minimize terms involving $z_i$ to reconcile the state $x_i$ with the dynamics propagated from the previous step. This forms a strictly convex least-squares problem, yielding the closed-form weighted average:
\begin{equation}
    z_i^{k+1} = \frac{\rho_{\text{eq}} \left( x_i^{k+1} + \rho_{\text{eq}}^{-1}\lambda_i^k \right) + \rho_{\text{dyn}} \left( d_{\text{dyn}, i-1}^{k+1} - \rho_{\text{dyn}}^{-1}\mu_i^k \right)}{\rho_{\text{eq}} + \rho_{\text{dyn}}}
\end{equation}

\textbf{Step 3: Geometric Layer Update $(\hat{x}, \hat{u})$.}
This step handles the physical inequality constraints. Taking $\hat{u}_i$ as an example, minimizing the Lagrangian reduces to evaluating a proximal operator \cite{parikh2014proximal}:
\begin{equation}
\hat{u}_i^{k+1}
= \operatorname*{argmin}_{\hat{u}}
I_{\mathcal{U}_i\cap\mathcal{U}_i^k}(\hat{u})
+ \frac{\rho_{\mathrm{geo}}}{2}
\biggl\|\hat{u}
-\bigl(u_i^{k+1}+\rho_{\mathrm{geo}}^{-1}\nu_{u,i}^k\bigr)
\biggr\|_2^2
\end{equation}
For standard box constraints (e.g., control limits and trust regions), this operation simplifies to an independent, element-wise clamping algorithm:
\begin{equation}
    \hat{u}_i^{k+1} = \text{Clamp}\left( u_i^{k+1} + \rho_{\text{geo}}^{-1}\nu_{u,i}^k, \ u_{\min}, \ u_{\max} \right)
\end{equation}
The update for the state proxy $\hat{x}_i$ is strictly analogous.

\textbf{Step 4: Dual Update.}
Finally, the dual multipliers are updated via parallel gradient ascent to penalize constraint violations, for example:
\begin{equation}
    \nu_{u,i}^{k+1} = \nu_{u,i}^k + \rho_{\text{geo}} (u_i^{k+1} - \hat{u}_i^{k+1})
\end{equation}
The four-step procedure is repeated until the primal and dual residuals satisfy the chosen convergence criteria.

\textbf{Remark on \ac{admm} Convergence:} Standard \ac{admm} guarantees convergence for two-block splitting. Our three-variable formulation maintains this guarantee via structural separability. Specifically, the variables can be grouped into two blocks: primal physical variables $(x, u)$ and auxiliary variables $(z, \hat{x}, \hat{u})$. Because the Augmented Lagrangian \cref{eq:augmented_lagrangian} lacks cross-penalty terms coupling the dynamic variables $z$ with the geometric mirror variables $(\hat{x}, \hat{u})$, their updates are conditionally independent given a fixed $(x, u)$. Consequently, Steps 2 and 3 evaluate in parallel as a single separable block update, reducing our framework to a standard two-block \ac{admm} sequence.

\textbf{Inexact inner solves:} Because \ac{admm} is a first-order method, achieving high precision demands significant computational time. Motivated by real-time requirements, we solve the inner \ac{admm} loop inexactly in our experiments rather than to strict tolerances. Consequently, we implement the method practically as an inexact SCP scheme.

\section{Numerical Experiments}
\label{sec:experiments}

To validate the proposed parallel-in-time solver, all experiments were executed on an NVIDIA Jetson AGX Orin 64GB embedded computing platform.
The solver is implemented in Python with a JAX/XLA backend, using \texttt{vmap} and \texttt{jit} to batch the per-node and per-scenario operations on the GPU.

\begin{figure}[h]
    \centering
    \includegraphics[width=0.5\linewidth]{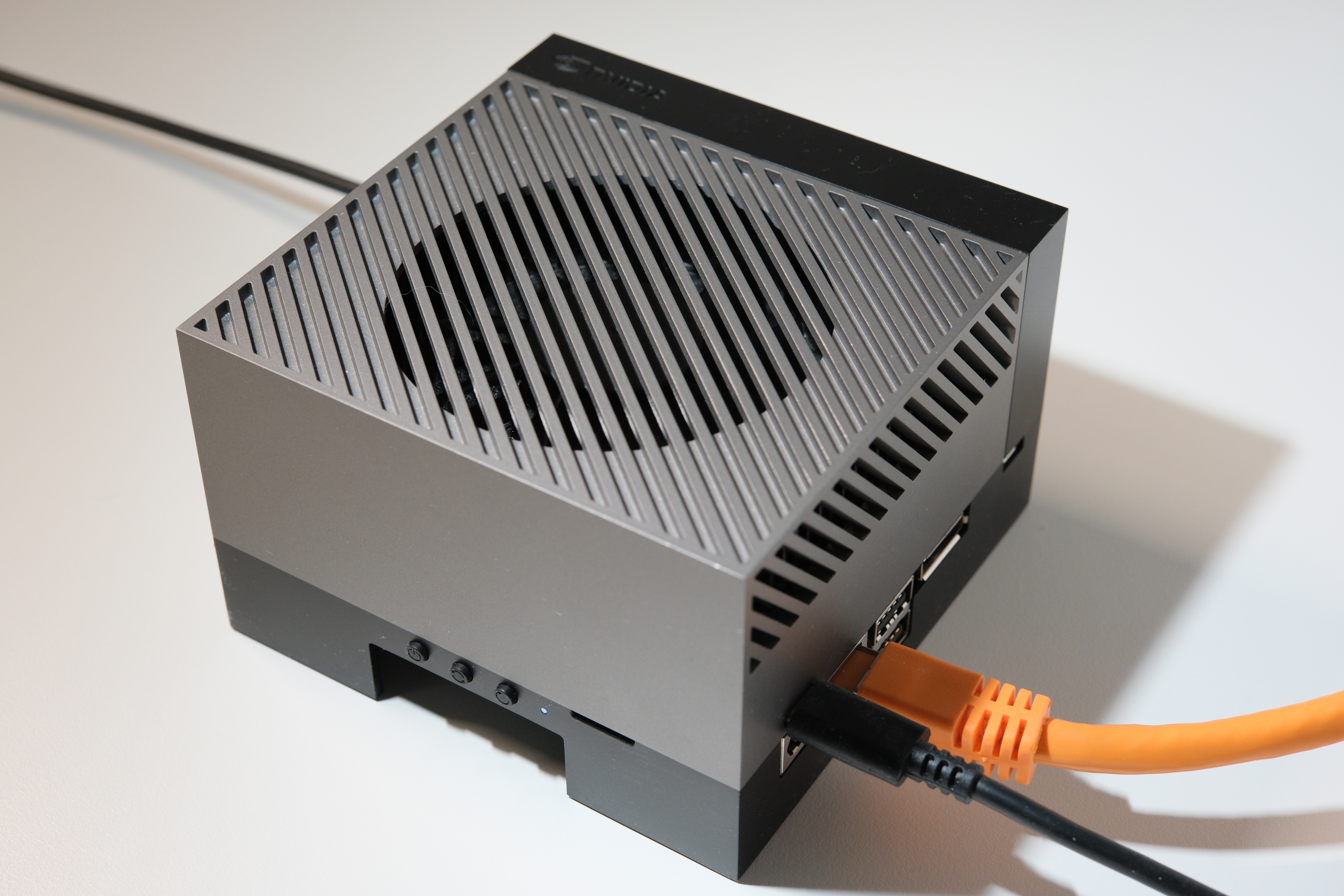}
    \caption{The NVIDIA Jetson AGX Orin 64GB edge computing platform used for hardware validation.}
    \label{fig:jetson_hardware}
\end{figure}

\subsection{Quadrotor Agile Flight}

\subsubsection{Problem Setup}
\label{subsubsec:quadrotor_setup}

We evaluate the solver on a 6-DOF quadrotor model. Agile quadrotor flight is a standard robotics benchmark due to its highly nonlinear, underactuated dynamics and the need for high-frequency replanning in cluttered environments \cite{foehn2021time}. Demonstrating real-time, massively parallel trajectory generation for this platform addresses a critical need for robust edge computing in autonomous navigation.

The system state $x \in \mathbb{R}^{13}$ comprises position $r \in \mathbb{R}^3$, linear velocity $v \in \mathbb{R}^3$, unit quaternion $q \in \mathbb{R}^4$, and body-frame angular velocity $\omega \in \mathbb{R}^3$. The control input $u \in \mathbb{R}^4$ consists of total thrust and body-frame torques.

The continuous-time dynamics follow standard Newton-Euler equations (detailed in \cite{folk2023rotorpypythonbasedmultirotorsimulator}), assuming a mass of $1.0\text{ kg}$ and gravity of $9.81\text{ m/s}^2$. These dynamics are discretized via a 4th-order Runge-Kutta scheme and linearized using JAX's automatic differentiation.

\subsubsection{Parameter Study}
\label{subsubsec:parameter_study}
For this study, the quadrotor is tasked with a 6.0~s flight ($N=50$) from an initial hover at $[-5.0, -5.0, 2.0]$~m to a target position at $[5.0, 5.0, 2.0]$~m. Terminal constraints strictly enforce the target position and zero linear velocity. The direct flight path is obstructed by three spherical obstacles with radii ranging from 1.2 to 1.5~m. Actuator limits constrain the total thrust to $[0, 20]$~N and body torques to $[-5, 5]$~Nm.

\begin{figure}[h]
    \centering
    \includegraphics[width=\linewidth]{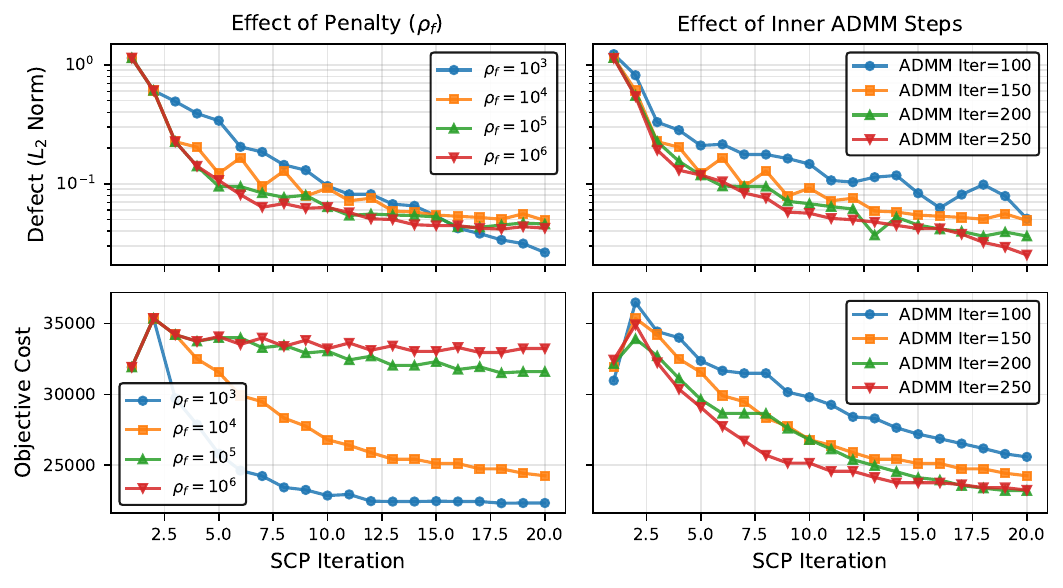}
    \caption{\textbf{Parameter Study.} Convergence profiles of the nonlinear dynamics defect (Top) and objective value (Bottom) for the quadrotor problem. All runs use $N=50$ and 20 \ac{scp} iterations. The left column varies the final penalty $\rho_f \in \{10^3, 10^4, 10^5, 10^6\}$ using a geometric schedule starting at $10^2$. The right column varies the fixed \ac{admm} inner-iteration count $\in \{100, 150, 200, 250\}$.}
    \label{fig:parameter_study}
\end{figure}

We evaluate the solver's sensitivity to the final penalty parameter $\rho_f$ and the number of inner \ac{admm} iterations under these tight conditions. For simplicity, the penalty parameters in \cref{eq:augmented_lagrangian} are set to an identical value ($\rho_{\text{eq}} = \rho_{\text{dyn}} = \rho_{\text{geo}} = \rho$) and scaled geometrically over 20 \ac{scp} iterations to a final value $\rho_f$. As shown in \cref{fig:parameter_study}, testing $\rho_f \in \{10^3, 10^4, 10^5, 10^6\}$ reveals a trade-off: higher values enforce constraint satisfaction more aggressively in early iterations but converge to higher final objective costs compared to $\rho_f=10^3$. Furthermore, the solver demonstrates robust inexact \ac{scp} behavior. It maintains stable convergence with as few as 100 inner \ac{admm} steps, though increasing the step count to 250 yields marginally lower final defects and objective costs.

\subsubsection{Scalability Benchmark}
\label{subsubsec:benchmark}
We evaluate the solver's robustness in randomized, obstacle-cluttered environments using a coarse temporal horizon ($N=40$). A batch of $B=1000$ noise-initialized trajectories is generated to navigate three random spherical obstacles (radii $1.0$ to $2.0$~m). Success is defined by three criteria: mean dynamics violation $< 10^{-2}$, obstacle penetration $< 10^{-3}$~m, and combined boundary condition error $< 0.1$. Under these thresholds, the solver achieves a 93.9\% success rate with an average dynamics defect of $0.007$.

\begin{figure}[h]
    \centering
    \includegraphics[width=\linewidth]{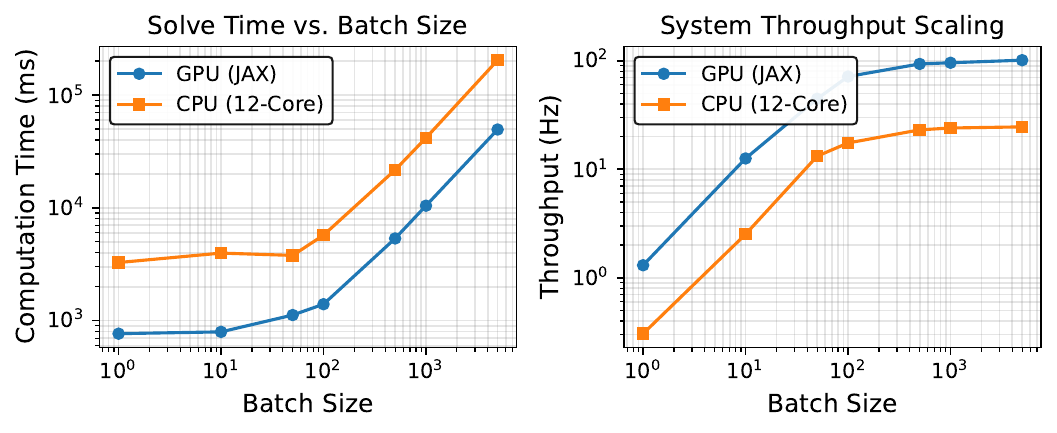}
    \caption{\textbf{Scalability Benchmark.} Computation time (Left) and throughput (Right) scaling comparison between the 12-core CPU baseline and the GPU solver.}
    \label{fig:benchmark}
\end{figure}

\begin{figure}[h]
    \centering
    \includegraphics[width=\linewidth]{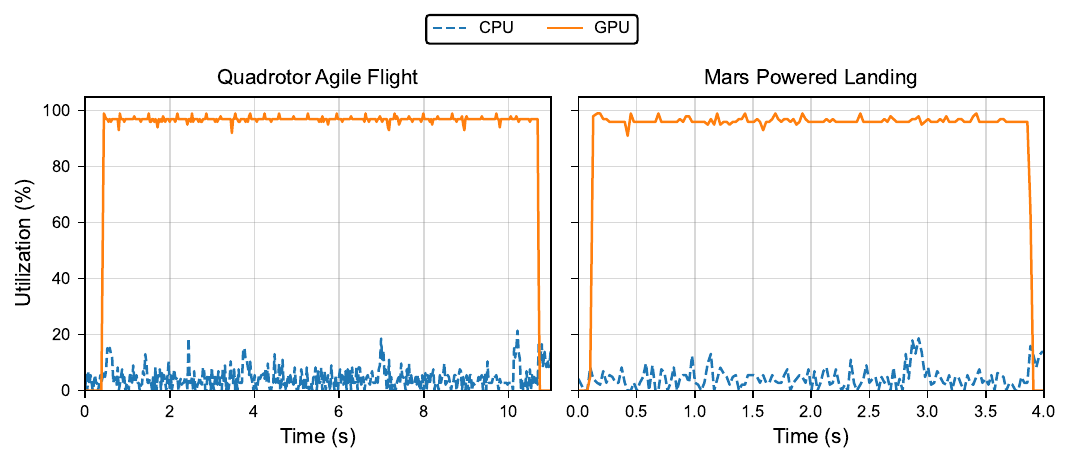}
    \caption{\textbf{Hardware Utilization Profiles.} Continuous profiling of CPU and GPU utilization during batched trajectory optimization. The solver maintains an active GPU utilization exceeding 96\% across both dynamical environments, effectively saturating the streaming multiprocessors.}
    \label{fig:utilization}
\end{figure}

To assess computational efficiency, we benchmark the GPU-native solver against a CPU baseline: a custom iLQR implementation featuring RK4 dynamics, finite-difference linearization, Levenberg-Marquardt regularization, clipped controls, and a backtracking line search. The baseline is accelerated via \texttt{numba} and parallelized across the Jetson AGX Orin's 12 CPU cores. \cref{tab:throughput} and \cref{fig:benchmark} illustrate scaling performance for batch sizes $B \in [1, 5000]$. The CPU throughput saturates near 24.6~Hz. Conversely, the GPU solver leverages JAX's \texttt{vmap} to reach a peak throughput of 101.1~Hz at $B=5000$, yielding a 4.1$\times$ speedup. Furthermore, the solver maintains 96.93\% active GPU utilization (\cref{fig:utilization}, left), effectively saturating the streaming multiprocessors.

This computational efficiency translates directly to energy savings. To quantify this, active energy consumption is calculated by trapezoidally integrating the total module power (\texttt{VIN\_SYS\_5V0} rail from \texttt{tegrastats} logs) over the active execution window. For a batch of $B=1000$ trajectories, the GPU solver consumes 119.03~J compared to 243.58~J for the multi-core CPU baseline, representing a 51\% reduction in energy expenditure. This efficiency offers a critical advantage for power-constrained mobile robots. Combined with a replanning rate exceeding 100~Hz, the solver is well-suited for real-time \ac{mpc}. In practical edge deployments, hardware power profiles and memory limits can be explicitly partitioned to reserve adequate compute and power budgets for concurrent sensor processing.

\begin{table}[h]
    \centering
    \caption{CPU vs. GPU Throughput and Speedup for Batched Trajectory Optimization}
    \label{tab:throughput}
    \begin{tabular}{cccc}
        \hline
        Batch ($B$) & CPU (Hz) & GPU (Hz) & Speedup \\
        \hline
        1 & 0.3 & 1.3 & 4.3$\times$ \\
        10 & 2.5 & 12.6 & 5.0$\times$ \\
        50 & 13.2 & 44.6 & 3.4$\times$ \\
        100 & 17.5 & 71.5 & 4.1$\times$ \\
        500 & 23.1 & 93.4 & 4.0$\times$ \\
        1000 & 24.0 & 95.7 & 4.0$\times$ \\
        5000 & 24.6 & 101.1 & 4.1$\times$ \\
        \hline
    \end{tabular}
\end{table}

\subsubsection{Robust Planning via Scenario Optimization}
\label{subsubsec:robust_tube}

We evaluate the solver's performance for robust \ac{mpc} via scenario optimization. The quadrotor navigates from $[-4.0, -5.0, 2.0]$ m to a target at $[5.0, 5.0, 2.0]$ m, bypassing a cylindrical obstacle with a 3.0 m radius. To ensure safety under uncertainty, we enforce an additional 0.5 m spatial buffer around the obstacle.

To simulate severe unmodeled disturbances, we inject a stochastic crosswind into the true dynamics within the middle flight corridor ($X \in (-2.5, 2.5)$). This wind applies a base acceleration of $[0.0, 2.5, -1.0]$ m/s$^2$ corrupted by Gaussian noise. At each \ac{mpc} step, a linear disturbance observer estimates the acceleration mismatch. To compute a safe policy, the solver jointly optimizes $K=15$ dynamically coupled scenarios over a horizon of $N=30$, where each scenario is injected with scaled noise to synthesize a distribution-aware safety tube. 

Crucially, because only one physical control can be applied to the robot, we employ a consensus-based approach \cite{bernardini2009scenario}. A non-anticipativity constraint restricts the first control input to be identical across all $K$ scenarios: $u^{(1)}_0 = u^{(2)}_0 = \dots = u^{(K)}_0$.
In our \ac{admm} implementation, this is imposed by replacing the first projected control of all scenarios with their scenario average at each inner iteration \cite{boyd2011distributed}. The \ac{mpc} controller applies this shared first control and warm-starts the subsequent solve by shifting the optimized scenario tube. This ensures the immediate action is physically executable while allowing predicted state trajectories to branch over the horizon to accommodate distinct disturbance profiles.

\begin{figure*}[t]
    \centering
    \includegraphics[width=0.65\linewidth]{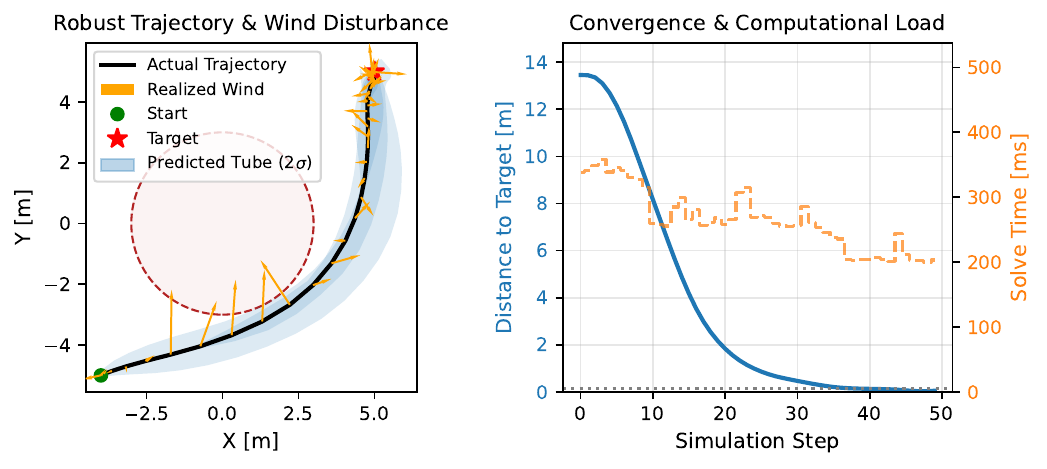}
    \caption{\textbf{Robust \ac{mpc} via Scenario Optimization.} (Left) Top-down trajectory plot illustrating the $2\sigma$ safety tube avoiding the obstacle under severe stochastic crosswind. (Right) Convergence of distance to the target and corresponding computation times per \ac{mpc} step.}
    \label{fig:robust_mpc}
\end{figure*}

As depicted in \cref{fig:robust_mpc}, the robust \ac{mpc} successfully anticipates the bounded uncertainty. It generates a $2\sigma$ spatial prediction tube that smoothly deforms to respect the safety margin despite the heavy crosswind. The quadrotor stabilizes at the target by step 40. Furthermore, the solver computes the full 15-scenario robust optimization in approximately 200 ms per step after stabilization, demonstrating its viability for complex, multi-trajectory constraints.

\subsection{Mars Powered Landing}

\subsubsection{Problem Setup}
\label{subsubsec:mars_setup}

We apply the proposed solver to the 6-DOF Mars powered descent landing problem adapted from Szmuk et al. \cite{szmuk2017successive}. In planetary landing scenarios, state estimation is frequently degraded by severe observation errors and sensor noise. Evaluating the extensive uncertainty bounds required for active guidance is computationally demanding. Our massively parallel approach addresses this challenge by enabling the simultaneous computation of hundreds of dispersed trajectories, facilitating real-time safety verification and robust control under high uncertainty. 

The problem is formulated using non-dimensional unified units. The state vector $x \in \mathbb{R}^{14}$ consists of vehicle mass $m$, inertial position $r \in \mathbb{R}^3$, velocity $v \in \mathbb{R}^3$, body attitude quaternion $q \in \mathbb{R}^4$, and angular velocity $\omega \in \mathbb{R}^3$. The control input $u \in \mathbb{R}^3$ is the thrust vector $T_B$ in the body frame.

The continuous-time dynamics account for mass depletion $\dot{m} = -\alpha \|T_B\|$ (where $\alpha = 0.1$), constant gravity, and full rotational kinematics governed by a diagonal inertia matrix. The planning horizon is fixed at 5.0 and discretized into $N=30$ intervals.

To ensure a safe and physically feasible descent, the vehicle must satisfy stringent state and control bounds. The spacecraft mass is restricted between a dry mass of 0.75 and a wet mass of 2.0. Actuator limits constrain the thrust magnitude within $[0.5, 3.0]$ and restrict the thrust gimbal angle to a $10^\circ$ cone relative to the body axis. Furthermore, spatial and attitude constraints are strictly enforced: the trajectory is confined within a $10^\circ$ glide slope cone to prevent surface collision, the vehicle tilt angle is limited to a maximum of $20^\circ$ from the vertical, and the angular velocity magnitude is bounded by $30^\circ$ per unit time. 

The spacecraft initiates the descent around a nominal position of $[2.0, 1.0, 0.0]$ in the Up-East-North frame with a velocity of $[-1.0, 0.2, 0.0]$. The solver is tasked strictly with minimizing fuel consumption to achieve a precise touchdown at the origin $[0.0, 0.0, 0.0]$ with a soft terminal velocity of $[-0.1, 0.0, 0.0]$ while maintaining a strictly upright orientation.

\subsubsection{Batch Trajectory Optimization and Scalability}
\label{subsubsec:mars_batch}

To evaluate the solver's computational throughput and its ability to guarantee safety under uncertainty, we conducted a massively parallel batch optimization experiment. To account for state estimation and observation errors, we applied a 5\% stochastic perturbation to the nominal initial state across all 14 state dimensions. By leveraging JAX's \texttt{vmap} transformation, the solver optimizes a batch of $B=1000$ independent descent scenarios in parallel. This large-scale Monte Carlo analysis serves to verify that, despite the initial dispersion, the solver can reliably ensure all trajectories remain within the feasible region and successfully reach the target.

\begin{figure*}[t]
    \centering
    \includegraphics[width=0.65\linewidth]{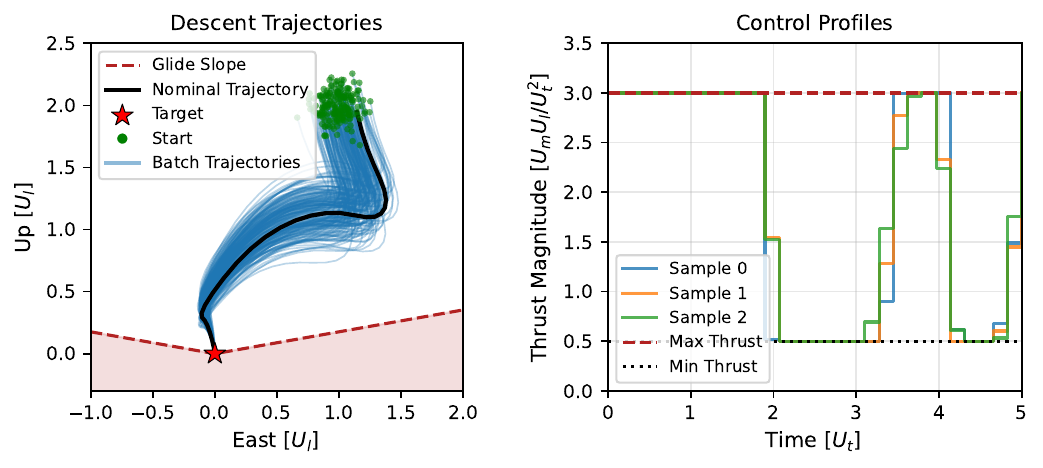}
    \caption{\textbf{Batch Mars Powered Descent.} (Left) 2D spatial bundle (Up vs. East) illustrating $200$ perturbed descent trajectories safely remaining above the strict glide slope boundary. (Right) Thrust magnitude profiles for selected samples, demonstrating the emergence of fuel-optimal bang-bang control characteristics.}
    \label{fig:mars_batch}
\end{figure*}

As depicted in \cref{fig:mars_batch}, the solver consistently discovers trajectories that remain within the $10^\circ$ glide slope boundary across the randomized initial conditions. Furthermore, the optimized thrust profiles exhibit notable bang-bang control characteristics, rapidly switching between maximum and minimum bounds to optimize fuel, although the structure is not strictly rigid due to the discrete-time formulation and algorithmic smoothing.

Under rigorous physical and algorithmic feasibility checks, including nonlinear dynamics rollouts, glide slope adherence, and boundary condition satisfaction, the solver achieved a success rate of 99.8\%, sustaining an overall computational throughput of 268.63 Hz. As demonstrated in the right panel of \cref{fig:utilization}, the solver achieved an active GPU utilization of 96.17\%, proving its ability to consistently saturate parallel hardware across distinct dynamic models. This performance demonstrates the solver's exceptional capability for rapid, large-scale trajectory generation and safety verification in complex aerospace applications.

\section{Conclusion}
\label{sec:conclusion}
This paper presented a GPU-native trajectory optimization framework that combines \ac{scp} with a consensus-based \ac{admm} decomposition to achieve massive parallelism across both time steps and problem instances. By splitting the nonlinear optimal control problem into independent per-node dense solves, closed-form dynamic consensus updates, and analytical constraint projections, the proposed method eliminates the need for global sparse factorizations, which often bottleneck conventional GPU implementations.

Numerical experiments on a 13-state quadrotor and a 14-state Mars powered descent problem validated the approach. The scalability benchmark demonstrated a sustained 4.1$\times$ throughput improvement and a 51\% reduction in energy consumption over a 12-core CPU baseline, achieving over 100 Hz replanning rates suitable for real-time \ac{mpc}. Furthermore, the framework's natural support for multi-trajectory optimization was demonstrated through scenario-based robust \ac{mpc}, where 15 dynamically coupled scenarios were jointly optimized under stochastic disturbances while respecting safety constraints. The solver's ability to efficiently handle complex, multi-trajectory constraints in realistic environments highlights its potential for deployment in real-world robotic systems, particularly those requiring high-frequency replanning and robustness to uncertainty.

Despite these computational advantages, several avenues remain for future work. First, an in-depth theoretical analysis is required to establish rigorous convergence guarantees, particularly concerning the interaction between the partially converged inner \ac{admm} loop and the outer \ac{scp} iterations. Second, future efforts will focus on deploying the GPU-native solver in physical hardware experiments to achieve real-world validations.

\section*{Acknowledgment}

We acknowledge the use of Gemini for English language checking and polishing throughout the text of this manuscript. The authors used this tool strictly to improve grammar, clarity, and readability. Following the AI assisted language refinement, the authors carefully reviewed and manually edited the entire manuscript to ensure factual accuracy, scientific correctness, and the integrity of the original ideas presented.

\bibliographystyle{IEEEtran}
\bibliography{ref.bib}

\end{document}

%% file: acronyms.tex
\DeclareAcronym{ode}{
  short = ODE,
  long  = Ordinary Differential Equation,
}

\DeclareAcronym{ocp}{
  short = OCP,
  long  = Optimal Control Problem,
}

\DeclareAcronym{nlp}{
  short = NLP,
  long  = Nonlinear Programming,
}

\DeclareAcronym{scp}{
  short = SCP,
  long  = Sequential Convex Programming,
}

\DeclareAcronym{admm}{
  short = ADMM,
  long  = Alternating Direction Method of Multipliers,
}

\DeclareAcronym{qp}{
  short = QP,
  long  = Quadratic Program,
}

\DeclareAcronym{ddp}{
  short = DDP,
  long  = Differential Dynamic Programming,
}

\DeclareAcronym{sqp}{
  short = SQP,
  long  = Sequential Quadratic Programming,
}

\DeclareAcronym{ipm}{
  short = IPM,
  long  = Interior Point Method,
}

\DeclareAcronym{mpc}{
  short = MPC,
  long  = Model Predictive Control,
}

\DeclareAcronym{kkt}{
  short = KKT,
  long  = Karush-Kuhn-Tucker,
}

\DeclareAcronym{simt}{
  short = SIMT,
  long  = Single Instruction Multiple Threads,
}

%% file: illustration.tex
\begin{figure*}[h!]
    \centering
    \begin{tikzpicture}[
        >=Stealth,
        font=\sffamily\small,
        scale=0.75,         %
        transform shape,    %
        scpbox/.style={rectangle, draw=blue!60, fill=blue!5, thick, rounded corners, minimum height=1.2cm, text width=3.5cm, align=center},
        varbox/.style={rectangle, draw=black!70, thick, minimum height=0.8cm, text width=2.8cm, align=center},
        geobox/.style={varbox, fill=green!10, draw=green!60!black},
        physbox/.style={varbox, fill=red!10, draw=red!60!black},
        dynbox/.style={varbox, fill=orange!10, draw=orange!60!black}
    ]

    \node[scpbox] (ocp) {\textbf{Nonlinear OCP}};
    \node[scpbox, right=1.5cm of ocp] (linearize) {\textbf{SCP Linearization}};
    \node[scpbox, right=1.5cm of linearize] (qp) {\textbf{Convex QP Subproblem}};

    \draw[->, thick] (ocp) -- (linearize);
    \draw[->, thick] (linearize) -- (qp);
    
    \draw[->, thick] (qp.north) -- ++(0,0.8) -| node[pos=0.25, below] {Update Nominal Trajectory $(\bar{x}^k, \bar{u}^k)$} (linearize.north);

    \node[physbox, below=3.8cm of linearize] (phys_i) {\textbf{Physical Layer} \\$x_i, u_i$};
    \node[geobox, above=0.8cm of phys_i] (geo_i) {\textbf{Geometric Layer} \\$\hat{x}_i, \hat{u}_i$};
    \node[dynbox, below=0.8cm of phys_i] (dyn_i) {\textbf{Dynamic Layer} \\$z_i$};
    
    \node[physbox, left=1.8cm of phys_i] (phys_im1) {\textbf{Physical Layer} \\$x_{i-1}, u_{i-1}$};
    \node[geobox, above=0.8cm of phys_im1] (geo_im1) {\textbf{Geometric Layer} \\$\hat{x}_{i-1}, \hat{u}_{i-1}$};
    \node[dynbox, below=0.8cm of phys_im1] (dyn_im1) {\textbf{Dynamic Layer} \\$z_{i-1}$};

    \node[physbox, right=1.8cm of phys_i] (phys_ip1) {\textbf{Physical Layer} \\$x_{i+1}, u_{i+1}$};
    \node[geobox, above=0.8cm of phys_ip1] (geo_ip1) {\textbf{Geometric Layer} \\$\hat{x}_{i+1}, \hat{u}_{i+1}$};
    \node[dynbox, below=0.8cm of phys_ip1] (dyn_ip1) {\textbf{Dynamic Layer} \\$z_{i+1}$};

    \foreach \step in {im1, i, ip1} {
        \draw[<->, thick, gray] (phys_\step) -- (geo_\step) node[midway, right, text=black, scale=0.8] {$\rho_{\text{geo}}$};
        \draw[<->, thick, gray] (phys_\step) -- (dyn_\step) node[midway, right, text=black, scale=0.8] {$\rho_{\text{eq}}$};
    }

    \draw[<->, thick, orange!80!black] (phys_im1.south east) -- (dyn_i.north west) node[midway, below left, scale=0.8] {$\rho_{\text{dyn}}$};
    \draw[<->, thick, orange!80!black] (phys_i.south east) -- (dyn_ip1.north west) node[midway, below left, scale=0.8] {$\rho_{\text{dyn}}$};

    \begin{scope}[on background layer]
        \coordinate (top_padding) at ([yshift=1.4cm]geo_i.north);
        \coordinate (bottom_padding) at ([yshift=-1.4cm]dyn_i.south);

        \node[fit=(geo_im1)(dyn_ip1)(top_padding)(bottom_padding), fill=gray!15, rounded corners, inner xsep=0.8cm, inner ysep=0cm] (gpu_bg) {};
        
        \node[above=0.2cm of gpu_bg.south, text=black] {\textbf{Massively Parallel GPU Execution}};

        \node[fit=(geo_im1)(dyn_im1), fill=white, draw=gray!60, dashed, thick, rounded corners, inner sep=0.3cm] (thread_im1) {};
        \node[above=0.1cm of thread_im1.north, text=gray!80!black] {\textbf{Thread Block $i-1$}};

        \node[fit=(geo_i)(dyn_i), fill=white, draw=gray!60, dashed, thick, rounded corners, inner sep=0.3cm] (thread_i) {};
        \node[above=0.1cm of thread_i.north, text=gray!80!black] {\textbf{Thread Block $i$}};

        \node[fit=(geo_ip1)(dyn_ip1), fill=white, draw=gray!60, dashed, thick, rounded corners, inner sep=0.3cm] (thread_ip1) {};
        \node[above=0.1cm of thread_ip1.north, text=gray!80!black] {\textbf{Thread Block $i+1$}};
    \end{scope}
    
    \draw[thick, gray!80, dashed] (qp.south west) -- (gpu_bg.north west);
    \draw[thick, gray!80, dashed] (qp.south east) -- (gpu_bg.north east);

    \end{tikzpicture}
    
    \caption{Hierarchical architecture of the proposed parallel-in-time trajectory optimizer.}
    \label{fig:method_overview}
\end{figure*}